\documentclass[11pt]{article}
\pdfoutput=1
\usepackage{acl}

\usepackage{times}
\usepackage{latexsym}

\usepackage{soul}

\usepackage{microtype}
\usepackage{amsmath,amsthm,amssymb}
\usepackage{mathtext}
\usepackage[T1]{fontenc}
\usepackage[utf8]{inputenc}
\usepackage[english]{babel}
\usepackage{graphicx}
\usepackage{hyperref}
\usepackage{upgreek}
\usepackage{todonotes}

\usepackage[utf8]{inputenc}

\title{SumHiS: Extractive Summarization Exploiting Hidden Structure}
\author{Pavel Tikhonov \\
ITMO University,\\
St. Petersburg, Russia\\
{\footnotesize \texttt{pav3370@yandex.ru}} \\\And
Anastasiya Ianina \\
MIPT,\\
Moscow, Russia\\
{\footnotesize \texttt{yanina@phystech.edu}}\\\And
Valentin Malykh \\
Kazan Federal University,\\
Kazan, Russia\\
{\footnotesize \texttt{valentin.malykh@phystech.edu}}\\}

\begin{document}
\maketitle

\begin{abstract}
Extractive summarization is a task of highlighting the most important parts of the text. We introduce a new approach to extractive summarization task using hidden clustering structure of the text. Experimental results on CNN/DailyMail demonstrate that our approach generates more accurate summaries than both extractive and abstractive methods, achieving state-of-the-art results in terms of ROUGE-2 metric exceeding the previous approaches by 10\%. Additionally, we show that hidden structure of the text could be interpreted as aspects.

\end{abstract}

\section{Introduction}
Summaries are important for processing huge amounts of information. A good summary should be concise, accurate and easy-to-read. However, there can be multiple variants of a perfect summary, the same idea can be conveyed with various words. Moreover, people may find different facts of the main importance, waiting for them to be present in the summary. Most automatic text summarization algorithms do not take into account different aspects of the initial texts, providing a semantically neutral interpretation. We aim to bridge the gap between summarization approaches and aspect mining. Thus, we investigate two research directions within this work: text summarization and aspect extraction.

\begin{figure}[tbh]
    \centering
    \includegraphics[width=\linewidth]{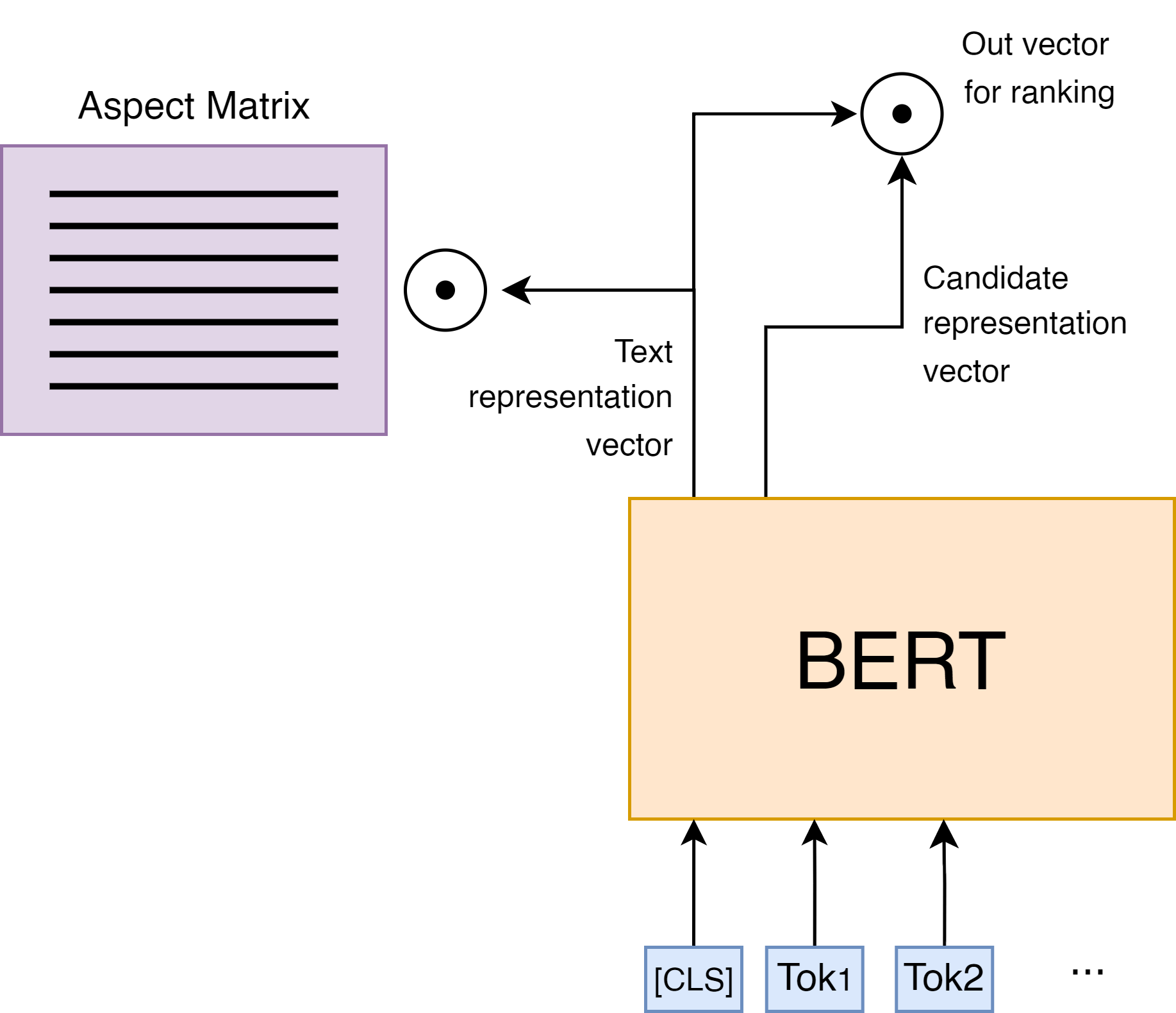}
    \caption{SumHiS: ranking model (right) + hidden structure discovery model (left).}
    \label{fig:model_general}
\end{figure}

\textbf{Text Summarization.}
There are two main approaches to text summarization: extrative and abstractive. Extractive methods highlight the most relevant phrases or sentences in the original text to form a summary. Alternatively, abstractive methods rephrase the text into a different form, and may not preserve the original semantic content.

Usually summarization has an underlying suggestion, that one summary should fulfill every informational demand. That is not true in many cases, e.g. imagine text about fruits in general, while a person is interested exactly in apples. In that toy example the proper summary for the aforementioned person should contain maximum information about apples with some occasional references to other fruits. Such a result can be achieved with aspect extraction techniques. The aspect extraction underlying suggestion is that each document consists of several aspects.

\textbf{Hidden Document Structure}.
Revealing hidden document structure is important for getting a concise and accurate summary. One way to do so is via aspect extraction. Each aspect may be specified by explicit words or sometimes inferred implicitly from the text.
For example, in the sentence ``the image is very clear'' the word “image” is an aspect term. The associated problem of aspect categorization is to group the same aspect expressions into a category. For example, the aspect terms ``image,'' ``photo,'' and ``picture'' can be grouped into one aspect category named Image. 

Hidden document structure is conventionally associated with dividing a document into multiple facets, each of which may have its own sentiment. However, the structures may relate to different textual features, e.g. topics covered in the text. In this paper we concentrate on how the discovered structures helps to make the summaries more accurate. Although we do not interpret these discovered structures as aspects.

\textbf{Our Approach.} We propose an extractive summarization model, that we call SumHiS (\textit{Sum}marization with \textit{Hi}dden \textit{S}tructure), which utilizes representations from BERT model~\cite{devlin2018bert} and uses topical hidden document structure. In this work we introduce two blocks for creating extractive summaries. First, we use contextualized representations retrieved from a pre-trained language models to rank the sentences from a document according to their importance. Second, we further filter the already ranked sentences in order to focus the summary on the facts corresponding to main discovered topics within document.

Evaluated on CNN/DailyMail dataset \cite{nallapati2016abstractive}, our approach outperforms previous extractive summarization state-of-the-art 
in terms of ROUGE-2~\cite{lin2004rouge} metric by 10\%. This results demonstrate the importance of topical structure inclusion for summarization task. 
Furthermore, we capitalize on the power of pre-trained language models combined with document structure discovery, that makes the resulting summary to focus on the most important topics and ideas mentioned in the initial text.

To summarize our key contributions are:
\begin{enumerate}
    \item A novel extractive summarization pipeline, which combines representations from pre-trained language models and hidden document structure discovery techniques.
    \item Our method outperforms prior work on the CNN-DailyMail dataset by a large margin in terms of ROUGE-2 and ROUGE-L metrics and can successfully be applied to real-world applications.
    \item Moreover, our model outperforms abstractive models too.
\end{enumerate}

The code of our system will be open-sourced shortly after the anonymity period.

\section{Related Work}
\label{sec:related}
The earliest attempts of automatic summarization focused on extractive techniques, which find words or sentences in a document that capture its most salient content.
Recent works use a variety of approaches. For example, \cite{zhong2020extractive} proposed a novel summary-level framework MatchSum and conceptualized extractive summarization as a semantic text matching problem. The authors proposed a Siamese-BERT architecture to compute the similarity between the source document and the candidate summary.
In \cite{dong2020hiporank} the authors rely on
extractive summarizers that identify salient sentences based on positional information.

Under supervised learning conditions, aspect-level sentiment classification is typically considered a classification problem. Early works \cite{boiy2009machine,kiritchenko2014nrc,wagner2014dcu} mainly used manually designed features such as sentiment lexicon, n-grams, and dependency information. However, these methods highly depend on the quality of the designed features,
which is labor-intensive. With the advances of deep learning methods, various neural models \cite{liu2017attention,chen2017recurrent,he2018exploiting} have been
proposed for automatically learning target-dependent sentence representations for classification. The
main idea behind these works is to develop neural architectures that are capable of learning continuous
features without feature engineering and at the same time capturing the intricate relatedness between a
target and context words.

Of course, there are many works in recent years in abstractive summarization. In the work \cite{nallapati2016abstractive} authors proposed to use encoder-decoder on a huge corpora to achieve good results in the abstractive summarization task.  Later in work \cite{nallapati2017summarunner} use a different type of recurrence network and obtained the state-of-the-art results. Nallapati and co-authors used copying word mechanism from the input sequence to the output, thereby solving the problem with rare words. In the paper \cite{cohan2018discourseaware} Cohan and co-authors proposed a summarization model for very long documents, like scientific articles. They use the hierarchical encoder mechanism that models the discourse structure of a document. Putra et al.~\cite{putra2018experiment} proposed to use so-called topical sentence, i.e. the most important one from the article, to generate news headline.

The last mentioned works allowed us to suggest a hidden structure usage in summarization. We chose a model which is designed to capture a hidden structure, namely extract aspects from texts.
Neural attention-based aspect extraction model (ABAE) is proposed in~\cite{he2017unsupervised}. The main idea of this work is to create a matrix of vector representations which could be used to reconstruct a sentence vector representation. It is done under assumption that there is only one main aspect which a sentence has.

In the models like MatchSum, the authors use vector BERT representations of the sentences. We decided to follow this approach, but instead of classic binary prediction whether a sentence should be included or not we chose ranking approach, allowing us to filter the sentences basing on their score.  
We chose recent state of the art approach in text ranking  SparTerm~\cite{bai2020sparterm}. This model is using vector representations of input texts to predict their ranking. The vector representations are produced from fine-tuned BERT model. We adopted this approach with exception of irrelevant to us term prediction task.

\section{Model Description}
This section presents the general overview of our extractive summarization system SumHiS, its architecture and the corresponding training strategy. Our system consist of two blocks: sentence ranking model and hidden structure discovery model. The models interaction is shown in Fig.~\ref{fig:model_general}. The training process of our system also consists of two phases. First, we train sentence ranking model and then we use its output representations to train a hidden structure discovery model.

\subsection{Ranking}
We follow Term-based Sparse representations (SparTerm) setup~\cite{bai2020sparterm} to train a ranking model. SparTerm learns sparse text representations by predicting the importance for each term in the vocabulary. Since we are working with extractive summarization, our model estimates importance of sentences instead of terms.

\begin{figure}[!tbh!]
    \centering
    \includegraphics[width=\linewidth]{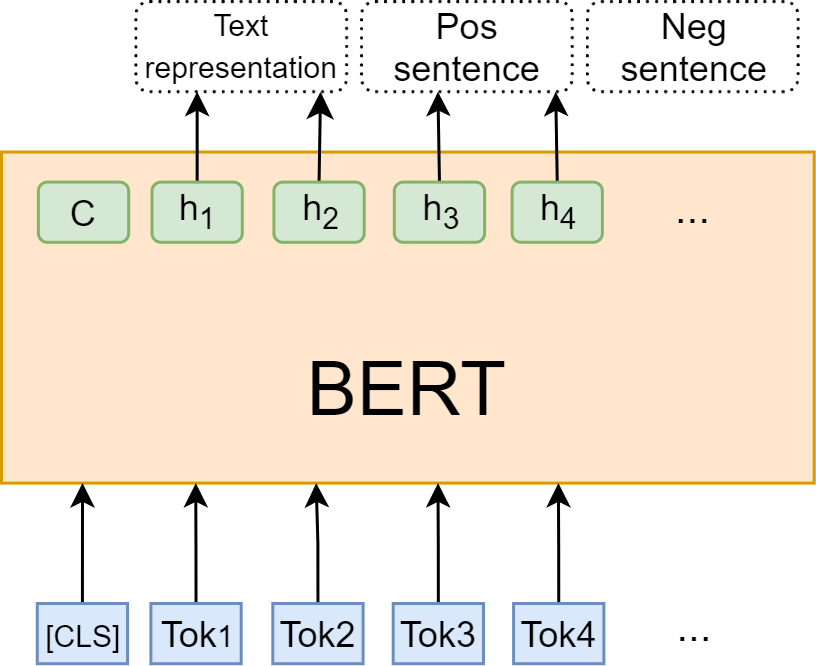}
    \caption{Ranking model}
    \label{fig:summarization_model}
\end{figure}

SparTerm represents text using BERT~\cite{devlin2018bert} model as follows: a text is fed into the model, and each term is embedded to a vector space. The term embeddings are averaged and used as a single text embedding. This text embedding is compared to other text embeddings thereby producing similarity scores.
Similarly to SparTerm setup, we use BERT with specifically designed input. 
Each input is represented as a triplet ($text$, $pos\_sentence$, $neg\_sentence$), where $text$ is a whole text of a document, $pos\_sentence$ is a sentence included into the golden summary, and $neg\_sentence$ is a sentence not included into the golden summary. 
The visualization of the model input is presented in Fig.~\ref{fig:summarization_model}.

We aim to make a positive sentence representation as close as possible to a representation of a text and simultaneously make a representation of negative sentence as far as possible from it. Let $ R=\{(t_{1},s_{1,+},s_{1,-}), ...,(t_{N},s_{N,+},s_{N,-})\} $ denote a set of $N$ training instances; each containing text $t_{i}$, positive candidate sentence $s_{i,+}$ and negative one $s_{i,-}$, indicating that $s_{i,+}$ is more relevant to the text than $s_{i,-}$. 
The ranking model is trained by optimizing a ranking objective which in our case is negative log likelihood of the positive sentence:
\begin{equation}
\begin{split}
L_{summ}(t_{i},s_{i,+},s_{i,-})= \\
-\log \frac{e^{sim(t_{i}^{'},s_{i,+}^{'})}}{e^{sim(t_{i}^{'},s_{i,+}^{'})}+e^{sim(t_{i}^{'},s_{i,-}^{'})}},
\label{eq:summ_loss}
\end{split}
\end{equation}
where $t_{i}^{'}$, $s_{i,+}^{'}$, $s_{i,-}^{'}$ are dense representations of $t_{i}$, $s_{i,+}$, $s_{i,-}$ respectively, and $sim$ denotes any similarity function. We use dot-product in our experiments.

During the training each document $t_{i}$ is split into sentences, from which the triplets are generated.
The output of the ranking model is an ordering of the sentences which are similar to the text summary from the closest one to the most distant one. Given this ordering one could create a summary for the text taking several top sentences.

\subsection{Hidden Document Structure Discovery}
Through experimentation, we found out that quality of summaries can be increased by adding information about hidden document structure.

We follow ABAE model \cite{he2017unsupervised} setup in order to capture hidden document structure. A sentence vector representation is considered to consist of weighted sum of several cluster representations. In case of ABAE these clusters are interpreted as aspects, while in our model we do not follow this interpretation and consider them as ordinal clusters.

The structure discovery model learns a matrix $C$ of cluster embeddings of size ${K \times n}$, where $K$ is the number of clusters, $n$ is an embedding space size. We use an attention-like mechanism to take into account all the cluster representations and reconstruct the initial vector. We calculate each input score by calculating its dot product with each cluster embedding:
\begin{equation}
{p}_{j} = c_j \cdot q
\end{equation}
where $q$ is an input text vector representation, while $c_j$ is $j$-th cluster embedding in the embedding matrix.
Obtained scores are then normalized with softmax function, leaving us with one highest weight corresponding to the leading cluster representation for $q$.
Next, each cluster vector is multiplied by the corresponding weight and summed up to get the output reconstructed vector $o$:
\begin{equation}
o = \sum\limits_{j=1}^K p_{j} c_j
\end{equation}

This output reconstructed vector is expected to be similar to the input text vector, so in order to train structure discovery model we minimize the loss function based on cosine distance:
\begin{equation}
L_{asp} = 1 - \frac{q \cdot o}{|q| |o|}
\label{eq:asp_loss}
\end{equation}

Such training allows us to build a model which could represent any input vector as a sum of one leading cluster representation and several others. This model is used for filtering of a vector set, it is an ordered set of sentence representations in our case. We filter the set in the following way. Let us say that $p_a^q$ is a weight for the leading cluster for the input text $q$. We could filter out any sentence $i$ from the set, where 
\begin{equation}
 p_a^i \le threshold,   
\end{equation}
where $threshold$ could be selected arbitrarily.

\section{Datasets}
\textbf{CNN/Daily Mail}~\cite{nallapati2016abstractive} is a dataset commonly used for text summarization evaluation. Human generated abstractive summary bullets were generated from news stories in CNN and Daily Mail websites as questions (with one of the entities hidden), and stories as the corresponding passages from which the system is expected to answer the fill-in-the-blank question. The authors released the scripts that crawl, extract and generate pairs of passages and questions from these websites.

All in all, the corpus has $286,817$ training pairs, $13,368$ validation pairs and $11,487$ test pairs, as defined by their scripts. The source documents in the training set have $766$ words spanning $29.74$ sentences on an average while the summaries consist of $53$ words and $3.72$ sentences.

\textbf{XSum}~\cite{narayan2018don}
is a dataset for evaluation of abstractive single-document summarization systems. The goal is to create a short, one-sentence new summary answering the question ``What is the article about?''. The dataset consists of 226,711 news articles accompanied with a one-sentence summary. The articles are collected from BBC articles (2010 to 2017) and cover a wide variety of domains (e.g., Politics, Sports, Weather, and Technology). The official split contains 204,045 (90\%), 11,332 (5\%) and 11,334 (5\%) documents in training, validation and test sets, respectively.

\subsection{Converting to Extractive Dataset}
Although the datasets are originally designed for abstractive summarization, we modified them for extractive summarization using a special utility. 
To obtain the extractive summaries from abstractive ones we use classic concept of extractive oracle. We define the extractive oracle summaries as follows, using ROUGE metrics described below:
\begin{equation}
\label{def:oracle}
\begin{split}
O= &~argmax_{S \subseteq D} \text{ROUGE}_N(G,S),\\
s.t.& ~~\ell(S) \le 2\ell(G). 
\end{split}
\end{equation}
Here $D$ is the set of all the sentences contained in the input document, and $G$ is the gold (abstractive) summary for the input document.
$\ell(\cdot)$ indicates the number of words in a text.

\begin{table*}[tbh!]
\centering
\begin{tabular}{|l|ccc|}
\hline
{Model}                     & {ROUGE-1} & {ROUGE-2}  & {ROUGE-L}  \\ \hline
MatchSum~\cite{zhong2020extractive} & \textbf{44.41}         & 20.86          & 40.55         \\
DiscoBERT~\cite{xu2020discourse} & \textit{43.77} &	20.85 &	\textit{40.67}         \\
BertSumExt~\cite{liu2019text}   & 43.85         & 20.34          & 39.90               \\\hline
SumHiS (w/o filtering) & 38.43         & \textit{28.51} & 37.58                \\
SumHiS (with filtering)    & 43.48         & \textbf{32.52} & \textbf{42.44} \\\hline    

\end{tabular}
\caption{ROUGE metrics for the extractive models on CNN/DailyMail test set (non-anonymized). Best result is given in bold, second best -- in italic.}
\label{tab:rouge_ext}
\end{table*}

\begin{table*}[tbh!]
\centering
\begin{tabular}{|l|ccc|}
\hline
{Model}                     & {ROUGE-1} & {ROUGE-2}  & {ROUGE-L}  \\ \hline
SimCLS~\cite{liu2021simcls} & \textbf{46.67} &	22.15 &	\textbf{43.54} \\
GSum~\cite{dou2021gsum}  & \textit{45.94} &	22.32 &	\textit{42.48} \\
Ours (w/o filtering) & 38.43         & \textit{28.51} & 37.58                \\
Ours (with filtering)    & 43.48         & \textbf{32.52} & 42.44 \\\hline    

\end{tabular}
\caption{ROUGE metrics for the abstractive \& our models on CNN/DailyMail test set (non-anonymized). Best result is given in bold, second best -- in italic.}
\label{tab:rouge_abs}
\end{table*}

\begin{table*}[tbh!]
\centering
\begin{tabular}{|l|ccc|}
\hline
{Model}                     & {ROUGE-1} & {ROUGE-2}  & {ROUGE-L}  \\ \hline
SimCLS~\cite{liu2021simcls} & \textbf{47.61} & 24.57 & 39.44 \\
GSum~\cite{dou2021gsum} & \textit{45.40} & 21.89 & 36.67 \\
\hline
MatchSum~\cite{zhong2020extractive} & 24.86 & 4.66 & 18.41 \\
BertSumExt~\cite{liu2019text} & 22.86 & 4.48 & 17.16 \\
\hline
Ours (w/o filtering) & 41.94         & \textit{28.80} & \textit{39.97}               \\
Ours (with filtering)    & 43.20         & \textbf{30.32} & \textbf{41.46} \\\hline    

\end{tabular}
\caption{ROUGE metrics for the XSum test set. Best result is given in bold, second best -- in italic.}
\label{tab:rouge_xsum}
\end{table*}

\section{Experiments}
\subsection{Metrics}
The models are evaluated with F1 variant (harmonic mean of Precision and Recall) of ROUGE-1, ROUGE-2, ROUGE-L~\cite{lin2004rouge}. ROUGE-N is computed as follows: 
\begin{center}
$ROUGE_N = \frac{\sum_{S\in Ref} \sum_{g_n\in S} Count_{match}(g_n)}{\sum_{S\in Ref} \sum_{g_n\in S} Count(g_n)}$
\end{center}

where $n$ stands for the length of the n-gram $g_n$, and $Count_{match}(g_n)$ is the maximum number of n-grams co-occurring in a candidate summary and a set of reference summaries $Ref$.

\begin{itemize}
\item ROUGE-1 value measures the overlap of unigram (each word) between the computed summary and the gold summary.

\item ROUGE-2 value measures the overlap of bigrams respectively.

 \item ROUGE-L measures the longest common subsequence between the model output and gold summary.

\item Recall in the context of ROUGE means how much of the gold summary is the computed summary capturing.

\item Precision answers how much of the computed summary was in fact relevant.

\end{itemize}

\subsection{Baselines}
We compare our model to the following models.
\paragraph{Extractive Models:}

\textbf{MatchSum}~\cite{zhong2020extractive}: this approach formulates the extractive summarization task as a semantic text matching problem. A good summary should be more semantically similar to the source document than the unqualified summaries.

\textbf{DiscoBERT}~\cite{xu2020discourse}: the model extracts sub-sentential discourse units (instead of sentences) as candidates for extractive selection on a finer granularity. To capture the long-range dependencies among discourse units, structural discourse graphs are constructed based on RST trees and coreference mentions, encoded with Graph Convolutional Networks.

\textbf{BertSumExt}~\cite{liu2019text}: the model uses pretrained BERT with inserted $[CLS]$ tokens at the start of each sentence to collect features for the sentence preceding it.

\paragraph{Abstractive Models:}
\textbf{SimCLS}~\cite{liu2021simcls}: a two-stage model for abstractive summarization, where a Seq2Seq model is first trained to generate candidate summaries with MLE loss, and then a parameterized evaluation model is trained to rank the generated candidates with contrastive learning.

\textbf{GSum}~\cite{dou2021gsum}: the model has two endoders which encode the source document and guidance signal, which are attended to by the decoder.

\textbf{ProphetNet}~\cite{qi2020prophetnet}: Transformer-based model which is optimized by n-step ahead prediction that predicts the next n tokens simultaneously based on previous context tokens at each time step.

\subsection{Experimental Setup}
For the summarization model, a pre-trained BERT was used (\emph{bert-base-uncased} variation from the \emph{Transformers} library~\cite{wolf-etal-2020-transformers}). Input sequence goes as follows: $$[CLS] text [SEP] sentence\_1 [SEP] 
sentence\_2$$  
$text$ is limited or padded to $430$ tokens, while $sentence\_1$ and $sentence\_2$ are both limited to $39$ tokens. $sentence\_1$ and $sentence\_2$ are filled with $pos\_sentence$ or $neg\_sentence$ randomly to force the model to not rely upon their relative ordering and use an embedded semantics.

During the evaluation, each document is split to sentences the exact same way as during the training. Each sentence is considered to be a candidate for inclusion in summary. It is fed into the model as $pos\_sentence$. As $neg\_sentence$ we use the last sentence in a text, since we assume it is never included into the summary. 

The structure discovery model is trained for two epochs, the threshold for filtering was set to $0.25$.

\section{Results}

We compared our model with current state of the art. We denote our model SumHiS with and without filtering for the variants of the model where the hidden structure discovery model is present or not respectively. 
We evaluate the models on the CNN/DailyMail dataset in non-anonymized version. The evaluation results are presented in Tab.~\ref{tab:rouge_ext}. One could see that our model shows the superior performance among the extractive models by the means of ROUGE-2 and ROUGE-L improving the previous results by almost 12\% and 2\% respectively. ROUGE-1 evaluation result for our model is 1 percent lower than state of the art result. Thus could conclude that our model is more successful in extraction of longer sequences of tokens, while keeping the unigrams distribution close to the desired one.

\begin{figure*}[tbh!]
    \centering
\fbox{\begin{minipage}{157mm}
\textit{Original text with highlighted extractive summary:}

MOSCOW, Russia ( CNN ). Russian space officials say the crew of the Soyuz space ship is resting after a rough ride back to Earth. \hl{A South Korean bioengineer was one of three people on board the Soyuz capsule.} \hl{The craft carrying South Korea 's first astronaut landed in northern Kazakhstan on Saturday, 260 miles ( 418 kilometers ) off its mark, they said.} Mission Control spokesman Valery Lyndin said the condition of the crew -- South Korean bioengineer Yi So - yeon, American astronaut Peggy Whitson and Russian flight engineer Yuri Malenchenko -- was satisfactory, though the three had been subjected to severe G - forces during the re - entry. Search helicopters took 25 minutes to find the capsule and determine that the crew was unharmed. Officials said the craft followed a very steep trajectory that subjects the crew to gravitational forces of up to 10 times those on Earth. Interfax reported that the spacecraft 's landing was rough. This is not the first time a spacecraft veered from its planned trajectory during landing. In October, the Soyuz capsule landed 70 kilometers from the planned area because of a damaged control cable. The capsule was carrying two Russian cosmonauts and the first Malaysian astronaut. 
\\
\\
\textit{Golden (abstractive) summary:}\\
Soyuz capsule lands hundreds of kilometers off-target.
Capsule was carrying South Korea's first astronaut.
Landing is second time Soyuz capsule has gone away.
\end{minipage}}
    \caption{Sample of SumHiS generated summary.}
    \label{fig:generated_summary}
\end{figure*}

In addition, we compare our model with abstractive models. The results are presented in Tab.~\ref{tab:rouge_abs}. Despite that our model is not using the generation, i.e. paraphrase ability of the language models, it shows the best results by ROUGE-2 metric outperforming the previous approaches by 10\%. ROUGE-L is evaluated only 1 percent lower that state of the art result. This result is an intriguing one, since the extracted bigrams are still better fit the desired distribution than the generated ones. 

It is important to mention, that structure discovery has significant influence on the model output, leading to improvement by 5\% in ROUGE-1 and ROUGE-L and by 4\% in ROUGE-2. We also provide a sample of SumHiS output in comparison to golden summary in Tab.~\ref{fig:generated_summary}.

\section{Analysis}

\begin{figure}[!tbh!]
    \centering
    \includegraphics[width=\linewidth]{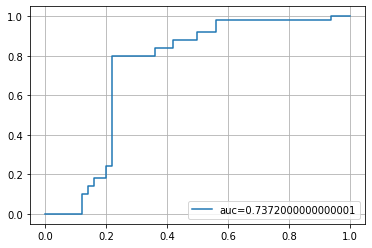}
    \caption{True Positive Rate vs. False Positive Rate for SumHiS with different threshold values.}
    \label{fig:roc_auc}
\end{figure}

The threshold in the experiments was not chosen randomly. We conducted a series of experiments resulting receiver output characteristic for the filtering classifier showed at Fig.~\ref{fig:roc_auc}). The vertical axis is true positive rate, while horizontal one is false positive rate. The value of $0.25$ shows the best balance between them.

\begin{table*}[tbh!]
\centering
\begin{tabular}{|l|ccccccccc|}
\hline
Model $\backslash$ {Metric} & R-1-p  & R-1-r & R-1-f & R-2-p  & R-2-r & R-2-f & R-L-p  & R-L-r & R-L-f \\\hline
SumHiS w/o filtering &   26.99&  79.01& 38.43 & 19.21&  65.59&  28.51 &   26.40 & 77.32 & 37.58 \\
SumHiS with filtering & 31.94 & 79.45& 43.49 & 22.77&  67.02& 32.52 & 31.16 & 77.62 & 42.44 \\
SumHiS + aspects & 19.24 & 100.00 & 31.12 & 13.67 & 98.64& 23.07 & 19.31 &100.00& 30.87 \\
SumHiS + binary loss  &  17.64&  100.00 & 28.61 & 12.47&  91.88 &  21.08 &  17.64& 100.00 &  28.61 \\
Orig. BERT  & 18.93 & 100.00 & 30.69 & 13.41 & 98.80 & 22.67 & 18.64 & 100.00 & 30.33\\
BERT with filtering & 19.12 & 99.99 & 31.01 & 13.59 & 98.77 & 23.02 & 19.08 & 99.99 & 30.51 \\
BERT + aspects & 18.44 & 92.12 & 29.33 & 12.75 & 78.03 & 21.11 & 18.32 & 91.53 & 29.15 \\\hline
\end{tabular}
\caption{Comparison of different variations of SumHiS model.}
\label{tab:ablation}
\end{table*}

\begin{table*}[tbh!]
\centering
\begin{tabular}{|l|c|}
\hline
\# & Aspect words \\\hline
1    & girl, women, teenagers, tennis, ladies, princess, kids                        \\
2    & perhaps, apparently, probably, mysteriously, presumably, suddenly, supposedly \\
3    & care, reasons, aids, purpose, irregularities, conform, attention              \\
4    & desperately, supplying, blood, terribly, abducted                             \\
5    & building, built, forcing, using, saying, trying                               \\\hline   
                                      
\end{tabular}
\caption{Sample of the extracted with SumHiS aspects.}
\label{tab:aspect_words}
\end{table*}

\subsection{Ablation Study}
The resulting SumHiS system has several choices which we did basing on the experiment results. SumHiS system contains two models, namely ranking and structure discovery ones. The choices for these models could be questioned thus we provide the results of an ablation study in Tab.~\ref{tab:ablation}. All the results are achieved on CNN/DailyMail dataset.
The metrics in this table are the variants of ROUGE, e.g. R-1-p is an abbreviation for ROUGE-1-Precision, R-2-r stands for ROUGE-2-Recall, while R-L-f means ROUGE-L F-measure variant of the metric. All the other metrics are named analogously.

We provide more complete results for \textit{SumHiS with} and \textit{without filtering}, naming them respectively in the table.
We have also tried to interpret the summarization task as binary classification problem, since it is a common approach in the field. In this setup we generate the following triplets: $(t_i, s_{i,+}, 1)$ and $(t_i, s_{i,-}, 0)$. The last value in a triplet is a label to predict. As a loss function we use classic binary cross-entropy. The results of this attempt are named ``\textit{SumHiS + binary loss}''. One could see that such replacement of a loss function is leading to catastrophic degradation of SumHiS quality by the means of Precision and F-measure as a consequence.

We considered the original BERT model without any fine-tuning on our data for the extractive summarization. We used the following setup as for SumHiS, we average per token representations to obtain the input text representation. To make an ordering required to produce a summary compare document text vector representation $t_i'$  with sentence representation $s_i'$. We use dot product as comparison function. The results for this model are denoted as \textit{Orig. BERT} in the table. Interestingly, Orig. BERT model shows better performance by ROUGE-2 (R-2-f in terms of Tab.~\ref{tab:ablation}), than BERT-based BertSumExt model, although the other metrics are significantly lower for it.

Next we applied our structure discovery model to the output of the Orig. BERT. The results are denoted as\textit{ BERT with filtering}. The structure discovery have improved all the metrics of Orig. BERT (not including R-1-r and R-L-r, since they were 100\%). The achieved result in ROUGE-2 (R-2-f) is a new state of the art, if we are leaving SumHiS aside. Although the improvement is small, it is consistent for all metrics. This result is correlated with filtering usage with SumHiS.

At last but not least we have experimented with our structure discovery model. It is partially following ABAE setup with two important differences: we do not used initialization for the clusters (aspects) and we do not regularize the cluster matrix. The initialization ABAE use is following: it takes vector representations of all the unique words in the training dataset; apply K-means clustering algorithm~\cite{Steinhaus1957} to the vectors where $K$ is set to be the desired number of aspects; and finally averaging all the vectors in a cluster to get its centroid vector. The centorids are used as initial values for the aspect embeddings.
The regularization which is used in ABAE is orthonormal one. It is formulated as follows:
\begin{equation}
    L_{ortho} = C \times C^T - I,
\end{equation}
where $C$ is an aspect matrix of size $K \times n$ and $I$ is diagonal unit matrix of size $K \times K$.
We have applied both of these techniques to our model and Orig. BERT. The results are denoted as ``\textit{+ aspects}'' in the table. Surprisingly to us, addition of aspect filtering is lowering all the metrics for both BERT and SumHiS models. Although the aspects extracted with this method seem to be adequate, the quality of the main task of summarization is too low to consider this approach as a general one. A sample of extracted with SumHiS aspects is presented in Tab.~\ref{tab:aspect_words}.

\subsection{Vector Space Analysis}
\begin{figure}[!tbh!]
    \centering
    \includegraphics[width=\linewidth]{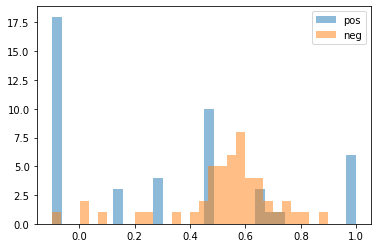}
    \caption{Histogram of distances between initial text and positive (blue) / negative (orange) sentences}
    \label{fig:pos_neg_comparison}
\end{figure}

\begin{figure}[!tbh!]
    \centering
    \includegraphics[width=\linewidth]{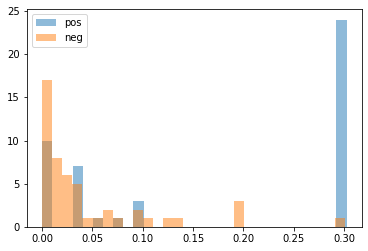}
    \caption{Histogram of distances between initial text and positive (blue) / negative (orange) sentences after kernel trick}
    \label{fig:kernel_trick}
\end{figure}

We aim the model to output different vectors for positive and negative input sentences.
To prove it, we calculated distances between $text'$ and $pos\_sentence'$ and $text'$ and $neg\_sentence'$ for every triplet in the test set.
As shown in Fig.~\ref{fig:pos_neg_comparison}, the distances between initial text representation and negative sentence representations are generally greater than the ones between the initial text and the positive sentence. To to take into account the peaking values we performed the kernel trick (see Fig.~\ref{fig:kernel_trick}): $(x-0.45)^2$, where $x$ is an initial distance between text and sentence.

\section{Conclusion}
We proposed a new model for extractive summarization that uses information about hidden document structure. Our model shows state-of-the art performance on CNN/DailyMail dataset by ROUGE-2 and ROUGE-L compared to current extractive summarization models. Moreover, it shows the best performance by the means of ROUGE-2 in comparison with abstractive models outperforming them by $10\%$. We showed that hidden structure in a text could be successfully used leading to significant improvements in summary generation.
As for the future work, we plan to make SumHiS end-to-end trainable aggregating ranking model and structure discovery models into an integral pipeline. We are also considering to integrate structure discovery within abstractive summarization, and experiment with different structure discovery mechanisms.

\bibliographystyle{acl_natbib}

\bibliography{lit}

\end{document}